\documentclass[hyphens]{article}
\usepackage[final]{neurips_2025} 

\usepackage[utf8]{inputenc} % allow utf-8 input
\usepackage[T1]{fontenc} % use 8-bit T1 fonts
\usepackage{hyperref} % hyperlink
\usepackage{subcaption}
\hypersetup{
 colorlinks=true,
 linkcolor=red,
 citecolor=cyan,
 filecolor=magenta, 
 urlcolor=magenta,
 }
\usepackage{url} % simple URL typesetting
\usepackage{booktabs} % professional-quality tables
\usepackage{amsfonts} % blackboard math symbols
\usepackage{nicefrac} % compact symbols for 1/2, etc.
\usepackage{microtype} % microtypography
\usepackage[dvipsnames,table]{xcolor}
\usepackage{amsmath}
\usepackage{xspace} 
\usepackage{enumitem} 
\usepackage{textcomp}
\usepackage{stfloats}
\usepackage{verbatim}
\usepackage{wrapfig}
\usepackage{graphicx}
\usepackage{subcaption}
\usepackage{amssymb}
\usepackage{cite}
\usepackage{tikz}
\usepackage{algorithm}
\usepackage{algpseudocode}

\usepackage{multicol,multirow} 
\usepackage{threeparttable} 
\usepackage{pgfplots}

\usepackage{fancyhdr}
\renewcommand{\headwidth}{\textwidth}

\renewcommand{\headrulewidth}{0.5pt}
\renewcommand{\headrule}{\vspace{2pt}\hbox to\headwidth{\color{black}\leaders\hrule height \headrulewidth\hfill}}
\pgfplotsset{compat=1.18}

\definecolor{totalbg}{gray}{0.92}
\def\modelname{X-Foresight}
\def\larchname{long-horizon chunk-wise auto-regressive strategy}
\def\vrname{Vision Renderer}
\def\ldmname{Large Drive Model}

\title{X-Foresight: A Joint Vision-Action Causal Forecasting Network via Predictive World Modeling}
\author{\colorbox{white}{\includegraphics[height=0.8em]{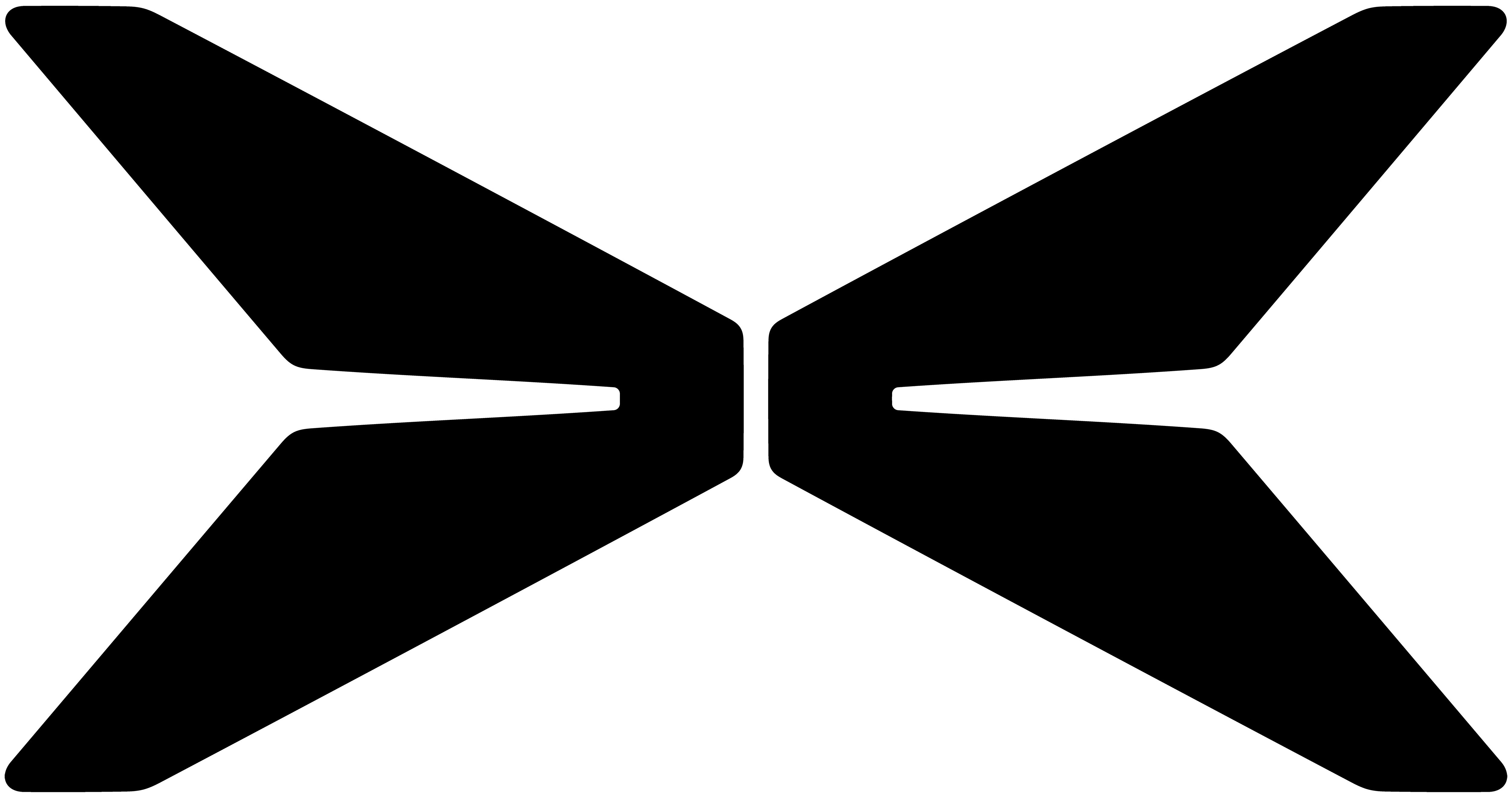}} PWM Team, XPeng Inc. \\
\textbf{\href{https://x-foresight-1.github.io}{\textcolor{NavyBlue}{https://x-foresight-1.github.io}}}}
\begin{document}
\maketitle

\begin{abstract}
Physical world knowledge resides mainly in videos. Equipping Vision-Language-Action (VLA) models with such knowledge is fundamental for safe and generalizable planning. To extract this knowledge from video data, predictive world modeling enables VLA to internalize physical dynamics and long-term causality by predicting future video from past observations. However, naive next-frame prediction faces two challenges: 1) unlike semantically distinct text tokens, video tokens are inherently low-entropy and redundant, causing prediction to easily degenerate into trivial extrapolation. 2) world modeling poses a temporal dilemma: instantaneous dynamics need dense frame prediction, while long-term causality unfolds over long and variable-size horizons that dense prediction cannot efficiently cover. To learn world knowledge effectively, we introduce \textbf{\modelname}, a predictive world model integrated directly into the VLA architecture to jointly learn world modeling and real-time action control. At its core lies \larchname{}, that addresses both challenges: By predicting across semantically distant chunks rather than adjacent frames, it escapes trivial extrapolation, while preserving dense intra-chunk frames to capture instantaneous dynamics and sparse inter-chunk transitions to capture long-term causality, resolving both the entropy and temporal-dilemma challenges at tractable training cost. A curriculum learning schedule is further utilized to progressively extend prediction horizons and stabilize long-horizon training. To capture long-term causality effectively, we present temporal importance sampling, which concentrates supervision on safety-critical chunks identified by ego-motion and behavioral signals. We further delegate photorealistic synthesis to a diffusion-based multi-view renderer, enriching photorealistic details and improving pixel-level appearance. Comprehensive experiments demonstrate that \modelname{} significantly outperforms VLA baselines in planning performance while maintaining strong generative fidelity, establishing a robust paradigm for world-knowledge-driven autonomous systems.
\end{abstract}

%-------------------------------------------------------------------------------------------
\begin{figure*}[h]
  \centering
   \includegraphics[width=\linewidth]{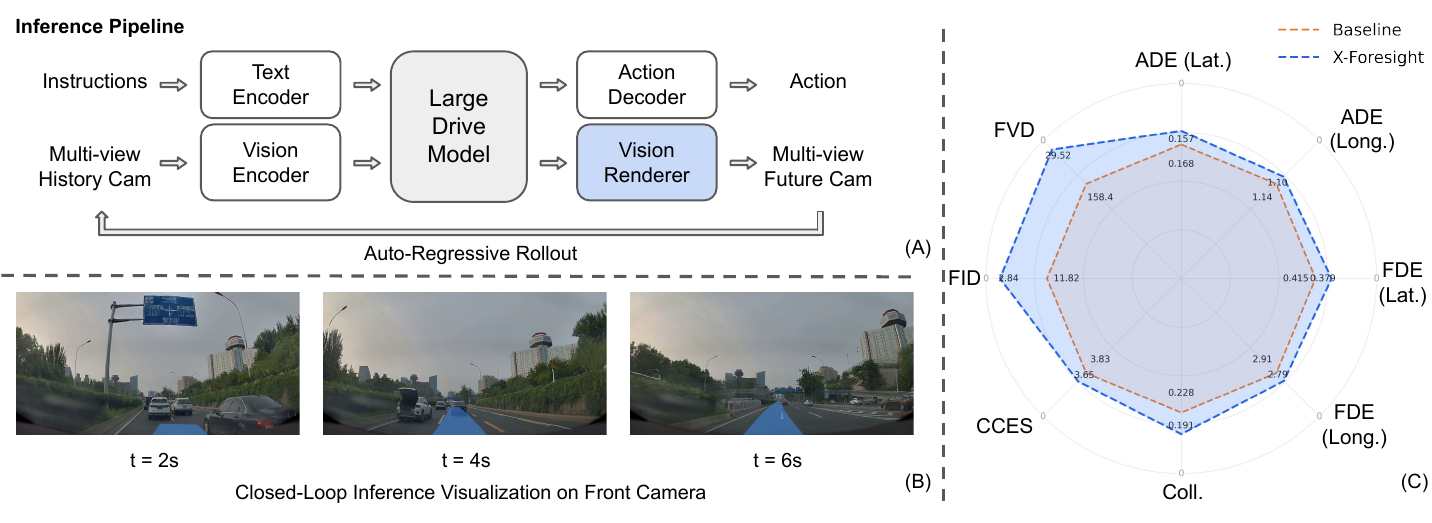}
   \caption{(A) Inference pipeline of X-Foresight. The main contributions resided in the Large Drive Model (LDM) and the Vision Renderer. (B) Closed-loop inference visualization of predicted future frames at t=2 s, t=4 s, and t=6 s; only the front camera was shown. (C) X-Foresight outperformed the baselines across multiple benchmarks.}
   \label{fig:main}
\end{figure*}
%-------------------------------------------------------------------------------------------

\section{Introduction}
The pursuit of artificial intelligence in physical world has driven significant interest in \textit{world models}. With world models, internal representations of the environment can be learned, which enforces agents' prediction, planning, and control. Recent world models employ diverse strategies to represent internal dynamics of the world: video-based generative models such as Sora~\cite{brooks2024video}, Genie~\cite{bruce2024genie}, and X-World~\cite{zheng2026x} learn to synthesize photorealistic future frames with large diffusion models; 3D-centric frameworks like Marble~\cite{world2026marble} explicitly model spatial geometry for physically consistent world reconstruction, while latent-based methods such as JEPA~\cite{assran2023self} bypass pixel-space generation entirely, learning semantic state representations through self-supervised predictive frameworks.

Vision-Language-Action (VLA) models have emerged as powerful paradigms for embodied decision-making, unifying perception, reasoning, and control within a single auto-regressive framework. Foundational works such as RT-2~\cite{zitkovich2023rt}, PaLM-E~\cite{driess2023palm}, and OpenVLA~\cite{kim2024openvla} have demonstrated remarkable generalization by transferring web-scale vision-language knowledge to robotic manipulation. In the autonomous driving domain, industry systems such as XPeng's VLA 2.0~\cite{xpeng2026vla2.0} exemplify the adoption of VLA architectures for end-to-end perception-to-control pipelines. However, prevailing VLA architectures remain fundamentally reactive, lacking deep world understanding and proactive danger-avoidance mechanisms. This limitation stems from their inability to simulate future states before acting, forcing policies to rely solely on historical observations. Without forward-looking foresight, models struggle to anticipate collisions, navigate through complex scenarios, or exploit long-horizon environmental causality to improve agent control.

Among the representations commonly used in robotics and autonomous driving, video provides one of the most holistic descriptions of the surrounding environment. It captures both low-level visual details, such as color, texture, and object shape, and high-level semantic information, such as the motion patterns of vehicles and vulnerable road users. Moreover, the ego vehicle's trajectory can often be inferred directly from dash-camera observations. Therefore, we argue that video serves as a primary carrier of physical-world knowledge, encoding rich spatial, temporal, and semantic cues that are essential for future prediction and decision-making. We thus introduce \modelname{}, a predictive world model built atop the VLA architecture that seamlessly integrates multi-view future prediction with closed-loop action control. By actively forecasting future camera images, the model develops an internalized understanding of long-horizon environmental causality, enabling proactive, safety-critical decision-making while preserving the real-time responsiveness required for autonomous systems.

Inspired by the success of large language models, future-token prediction has become a central paradigm for generative sequence modeling. However, directly transferring this paradigm to video-based world modeling introduces an important modality gap. Language tokens are semantically discrete, relatively sparse, and high-entropy, whereas video tokens are often redundant, with high frame-to-frame similarity. This difference can easily lead to degenerate future rollouts, where the model learns only trivial frame-level extrapolation rather than meaningful physical dynamics.
A further temporal challenge in world modeling arises from the mismatch between instantaneous dynamics and world transitions: effective world representation learning must capture short-term dynamics and long-term causality simultaneously. Instantaneous dynamics rely primarily on short-term dense visual evidence, whereas world transitions demand long-horizon causal reasoning over extended temporal contexts. To reconcile this discrepancy, \modelname{} employs a chunk-wise auto-regressive rollout that jointly forecasts future vision and action tokens. The proposed \larchname{} not only avoids trivial extrapolation by preserving dense intra-chunk frames that capture instantaneous dynamics, but also enables efficient training and low-latency inference.

We further improve world-knowledge learning through targeted training innovations. Since predictive capability develops progressively, we adopt a curriculum learning schedule in which the model first learns to forecast short horizons within each chunk and is then gradually exposed to longer prediction horizons. This progressive training strategy improves the stability of long-horizon forecasting and enhances the robustness of the learned policy.
Additionally, because temporal segments do not contribute equally to world modeling, we introduce a hybrid sampling strategy that combines random frame selection with importance-weighted sampling. This approach encourages broad temporal coverage while emphasizing safety-critical transitions, thereby improving the model's ability to learn informative world knowledge relevant to long-term causality and decision-making.
We argue that world representations should remain highly abstract; excessive visual detail at the latent level dilutes the model's capacity for structural world understanding. To reconcile abstraction with perceptual fidelity, we delegate photorealistic synthesis to a dedicated diffusion module. This enables high-fidelity multi-view rendering, where a diffusion-based renderer reconstructs detailed surround-view cameras within the auto-regressive loop, providing dense latent-space supervision.

In summary, the key contributions of this work are:
\begin{itemize}
\item We introduce a \larchname{} that leverages extended future horizons for world modeling. This design both mitigates the collapse of world-knowledge learning under naive next-frame prediction and resolves the temporal dilemma: dense intra-chunk frames allow the model to capture instantaneous dynamics, while long-horizon chunk-level prediction promotes the learning of broader world causality.
\item To improve long-horizon forecasting stability and enhance policy robustness, we exploit a curriculum-based learning strategy, starting with short-horizon prediction across chunks and gradually extending to longer strides.
\item We adopt temporal importance sampling, a hybrid sampling mechanism that accounts for uneven contributions of future frames to world knowledge learning, combining random selection with importance-weighted focus on critical temporal transitions.
\item The proposed method can acquire high-fidelity multi-view future images. Integration of a diffusion-based renderer into the auto-regressive pipeline is developed to reconstruct photorealistic surround-view details.
\end{itemize}

Comprehensive experiments demonstrate that \modelname{} significantly outperforms the reactive VLA baseline in planning safety, generative fidelity as shown in Fig~\ref{fig:main}, establishing a robust paradigm for world-knowledge-driven autonomous systems.
\section{Data}
\subsection{Data scale/resolution}
\begin{figure}[h]
    \centering
    \includegraphics[width=0.75\linewidth]{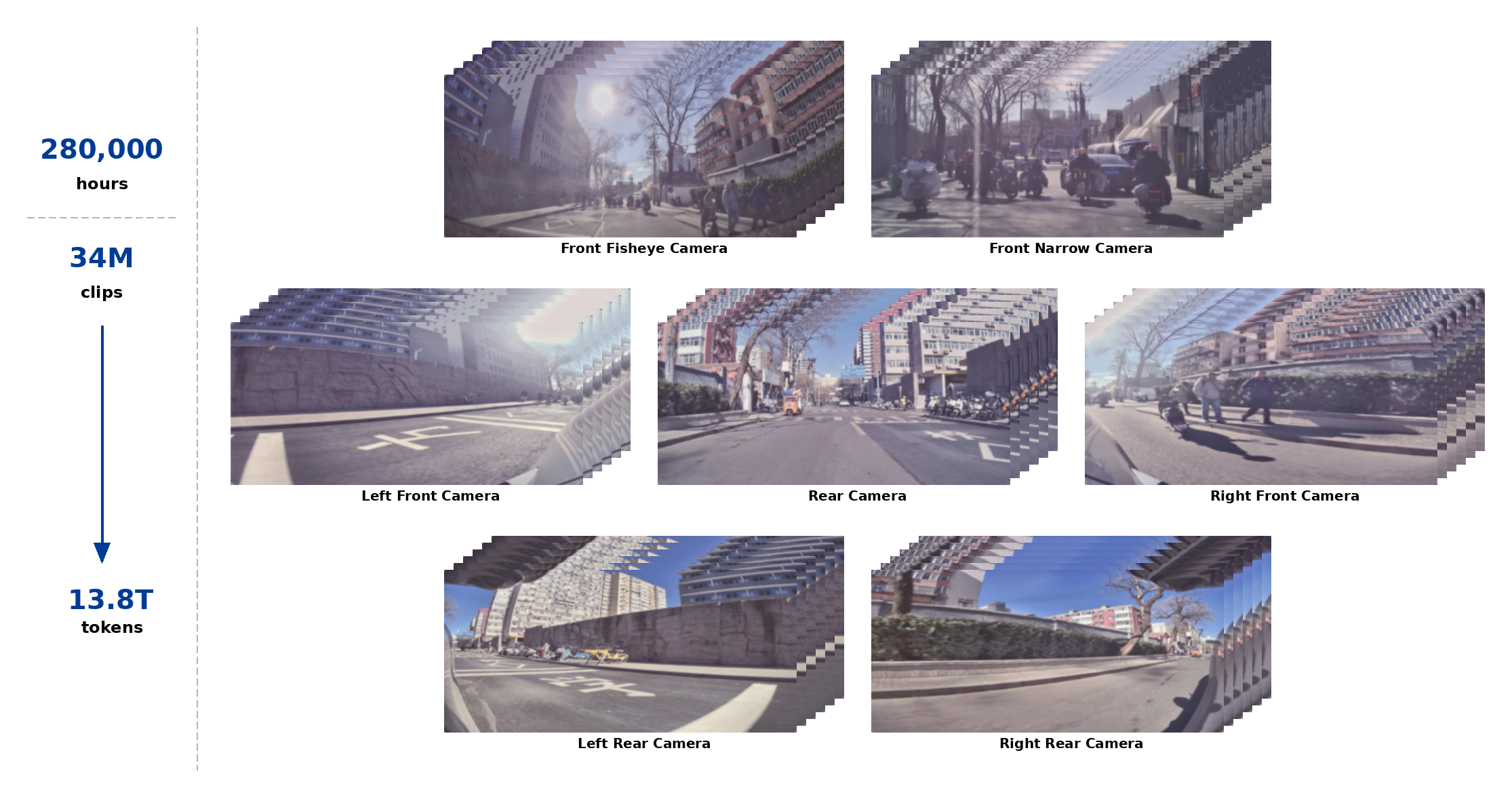}
    \caption{Overview of the large-scale multi-camera driving dataset. The dataset contained approximately \textbf{280,000 hours} of driving data, segmented into \textbf{34M clips}. Multi-view images from \textbf{seven cameras} (front fisheye, front narrow, left/right front, left/right rear, and rear) were processed and tokenized into \textbf{13.8T tokens}, which were used for training the world model.}
    \label{fig:dataset overview}
\end{figure}

  We built an industry-level dataset with approximately 280,000 hours of in-house driving data, spanning
  various traffic conditions and environmental scenarios, as illustrated in Fig~\ref{fig:dataset overview}.
  In total, the data was segmented into 34M clips of up to 30 seconds each and tokenized into 13.8T tokens
  from multi-view observations. All camera streams were stored at a native frequency of 12~Hz and
  downsampled to 4~Hz for training. This setup provided sufficient temporal context for long-horizon
  world modeling while maintaining a tractable sequence length for efficient training.

The dataset contained two native resolution configurations. 
To balance computational efficiency and spatial fidelity, images from different cameras were resized to viewpoint-specific resolutions to better preserve spatial characteristics. This temporal configuration provided a favorable trade-off between motion fidelity and computational cost, allowing the model to capture fine-grained dynamics while keeping sequence lengths tractable for training.

Data was collected using a 7-camera surround-view system that provides full 360-degree coverage. The setup consisted of one front-facing fisheye camera for wide field-of-view perception, one front-facing narrow camera for long-range observation, four side cameras for cross-traffic awareness, and one rear camera. The camera layout and mounting configuration were consistent with X-World~\cite{zheng2026x}. All cameras were geometrically calibrated with respect to the ego vehicle frame. This multi-camera configuration enabled consistent multi-view observations and reduced occlusions, which was critical for learning a spatially coherent and temporally consistent world model.

\subsection{Distribution}

\begin{figure}[t]
    \centering
    \includegraphics[width=0.92\linewidth]{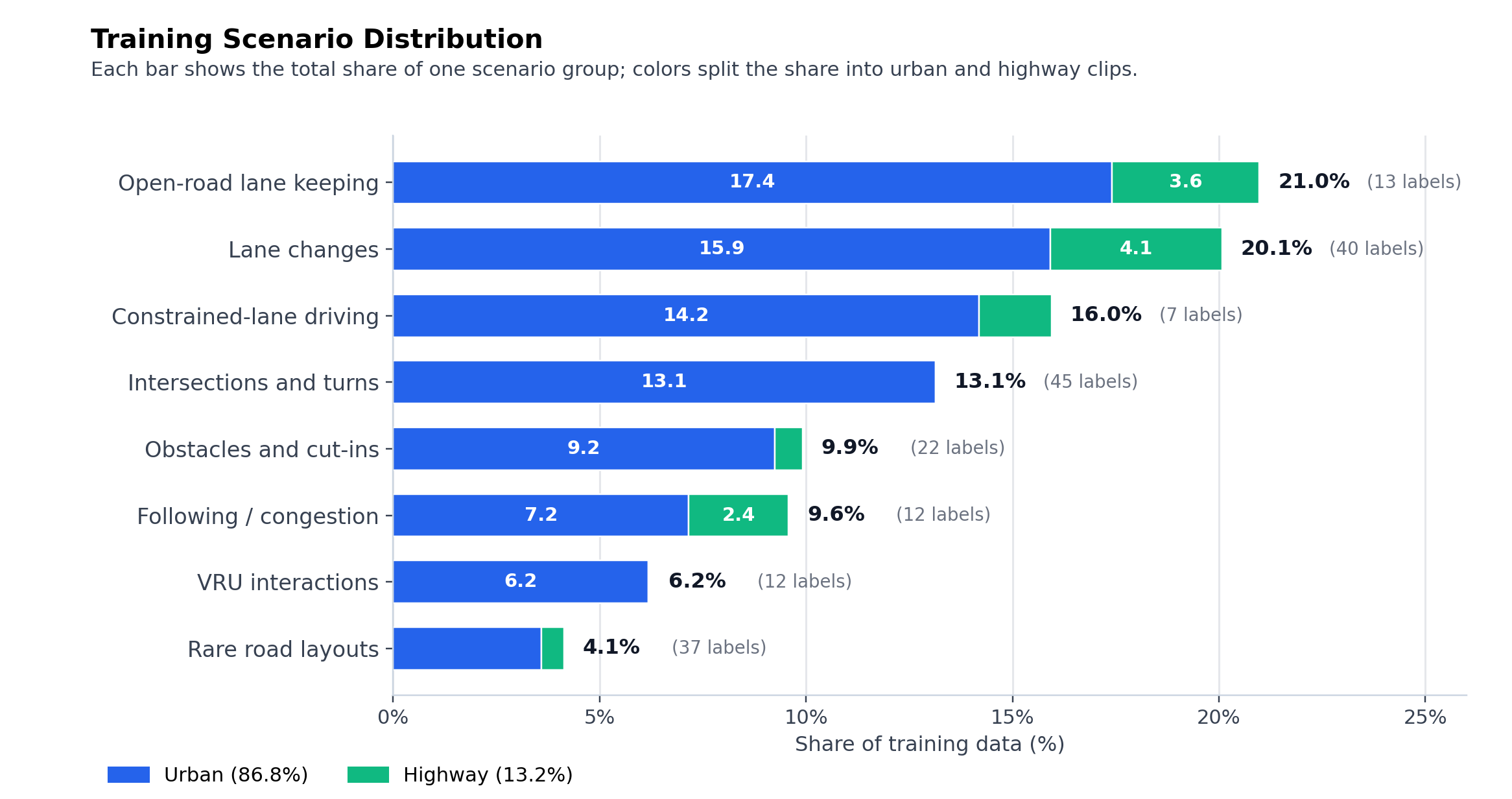}
    \caption{Training scenario distribution of \modelname. Fine-grained
    auto-tags were grouped into eight scenario categories. Bar length denoted
    the total share of the category, colors split the share by road class
    (urban / highway), and numbers in parentheses denoted the number of
    fine-grained labels assigned to each category.}
    \label{fig:data_distribution}
\end{figure}

Following the auto-tagging and distribution-analysis protocol of
X-World~\cite{zheng2026x}, we analyzed the scenario distribution of the
\modelname{} training set from auto-tagged driving clips. The original tags
covered nearly 200 fine-grained scenarios, including road type, ego maneuver,
road layout, surrounding traffic, and long-tail safety cases. Here, we
aggregated these fine-grained tags into eight scenario categories according
to the main driving context of each tag. This grouping preserved the main
driving regimes in the dataset while retaining explicit categories for
interaction-heavy and long-tail scenarios.
 
Fig~\ref{fig:data_distribution} showed the aggregated distribution.
Routine driving still took the largest share: open-road lane keeping
accounted for 21.0\% of the data and lane changes accounted for 20.1\%.
Constrained-lane driving, such as line straddling and unmarked roads, made
up 16.0\%, while intersections and turns contributed 13.1\%. The rest of the
data covered interaction and long-tail scenarios,
including obstacles and cut-ins (9.9\%), following and congestion (9.6\%),
VRU interactions (6.2\%), and rare road layouts such as ramps, toll
stations, roundabouts, auxiliary roads, and high-curvature segments (4.1\%).
Together, common driving and maneuvering cases accounted for about 70\% of
the training data, while the remaining portion provided coverage for more
challenging agent interactions and road structures.
 
The road-class split further showed that the dataset was dominated by urban
driving (86.8\%), with highway clips accounting for 13.2\%. This mix reflected
the operating regimes most frequently encountered in large-scale public-road
driving, while still preserving explicit coverage of safety-critical and
long-tail scenarios such as VRU interactions, cut-ins, ramps, and toll
stations. The resulting distribution provided broad supervision for
world-model learning and supported scenario-aware evaluation of \modelname{}
across both common driving patterns and challenging edge cases.

\section{Methodology}
\begin{figure*}[t]
  \centering
   \includegraphics[width=1.0\linewidth]{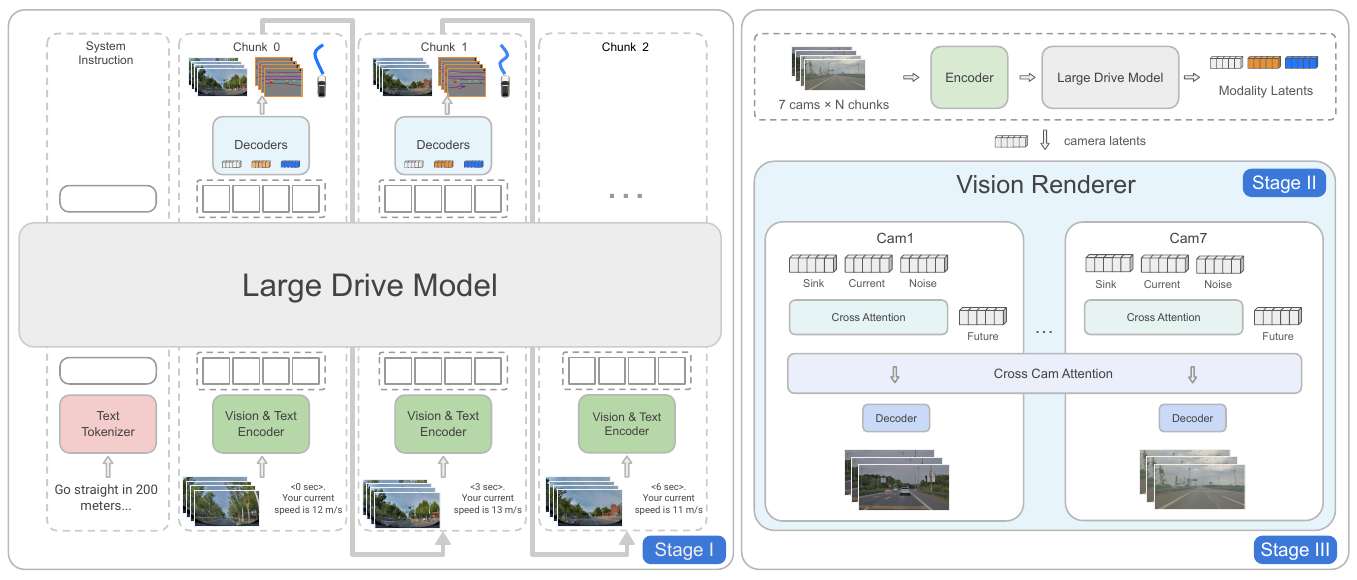}
   \caption{The training pipelines of the proposed \modelname{}.
\modelname{} consisted of two main components, a \ldmname{} (LDM) that performed joint world prediction and action planning in a unified token space, and a Vision Renderer that decoded the LDM's predicted camera tokens into photorealistic multi-view frames. In stage~I and stage~II, the LDM and the Vision Renderer were trained separately, with the renderer conditioned on ground-truth future trajectories; in stage~III, the LDM was frozen and the renderer's conditioning was swapped to the LDM's predicted future camera tokens.}
   \label{fig:train_pipeline}
\end{figure*}

This section presented the methods employed in \modelname{}.  \modelname{} consisted of two main components: a \ldmname{} (LDM) that performed joint world prediction and action planning in a unified token space, and a \vrname{} that decoded the LDM's predicted camera tokens into photorealistic multi-view frames as shown in Fig~\ref{fig:train_pipeline}. The LDM was an auto-regressive transformer that consumed a temporal stream of multi-view camera observations, language instructions, ego-vehicle state, and predicted three kinds of future targets at every step: ego actions used for real-time control, a bird's-eye-view (BEV) plot that captured the surrounding spatial structure, and per-camera latent tokens that summarized the predicted future appearance from each viewpoint. The Vision Renderer was a diffusion-based decoder whose rendered frames were fed back into the LDM as fresh observations, closing the auto-regressive prediction loop. We further described the training strategy and the inference pipeline of the proposed \modelname{}.
\subsection{\ldmname}

\subsubsection{Design Rationale}
To enable world understanding in \modelname{}, \ldmname{} (LDM) was explicitly trained to predict future images, bird's-eye-view (BEV) representations, and actions. We adopted a \larchname{} to perform future rollouts over extended horizons. Training on long-horizon data allowed the model to capture world transition causality while improving data utilization. Further details were provided in Section \ref{subsubsec:Extension to Longer Horizon}.

The chunk-wise design addressed an inherent temporal dilemma: instantaneous dynamics required densely sampled short-range frames, whereas long-term causality unfolded over variable-length horizons. Chunk-wise future rollout served as the key mechanism to reconcile these two requirements. Inspired by the success of autoregressive modeling in LLMs and VLMs, we adopted an autoregressive formulation that further enabled extensible future rollout generation.

The \larchname{} within LDM also facilitated rapid deployment. Since world knowledge was encoded during training, LDM at inference time was capable of both producing high-frequency instantaneous control signals and generating long-horizon rollouts with high-fidelity multi-view image synthesis, with the support of \vrname{}.

\subsubsection{Multi-modal Prompt Design}

The core component of the \modelname{} framework was the LDM trained to predict future ego actions and future observations simultaneously. 
As illustrated in Fig~\ref{fig:train_pipeline}, we formulated this problem as a multi-modal prompting task:
\begin{equation*}
    \space [\texttt{SYSTEM PROMPT}] \; | \; [l_{0},O_{0},A_{0},Q_{0}] \; | \; [l_1, O_1, A_1 ,Q_{1}] \; | \; \ldots \; | \; [l_{i}, O_{i}, A_{i},Q_{i}] \, .
\end{equation*}
The prompt consisted of a global system prompt followed by a sequence of temporal chunks. The system prompt provided general task instructions and high-level context about the ego vehicle, including its long-horizon navigation objective. Each temporal chunk contained four types of tokens. The text tokens $l_i$ specified the prediction horizon or temporal window. The observation tokens $O_i$ represented multi-camera video tokens extracted from multiple frames within the chunk using a ViT~\cite{dosovitskiy2020image} encoder. The action/state tokens $A_i$ encoded the ego-vehicle states, which can be derived from historical trajectories. The query tokens $Q_i$ were used to trigger the prediction of future variables. To reduce the overall token budget, video tokens were predicted in place rather than through additional visual query tokens.

\subsubsection{Extension to Longer Horizon}\label{subsubsec:Extension to Longer Horizon}
While our prompt design could naturally extend to longer horizons, directly increasing the prediction duration was computationally prohibitive: both memory footprint and computational cost grew quadratically with sequence length. In this section, we introduced several key designs which not only improved model performances, but also optimized LDM training on longer horizon.

\begin{figure}[h]
    \centering
    \includegraphics[width=0.9\textwidth]{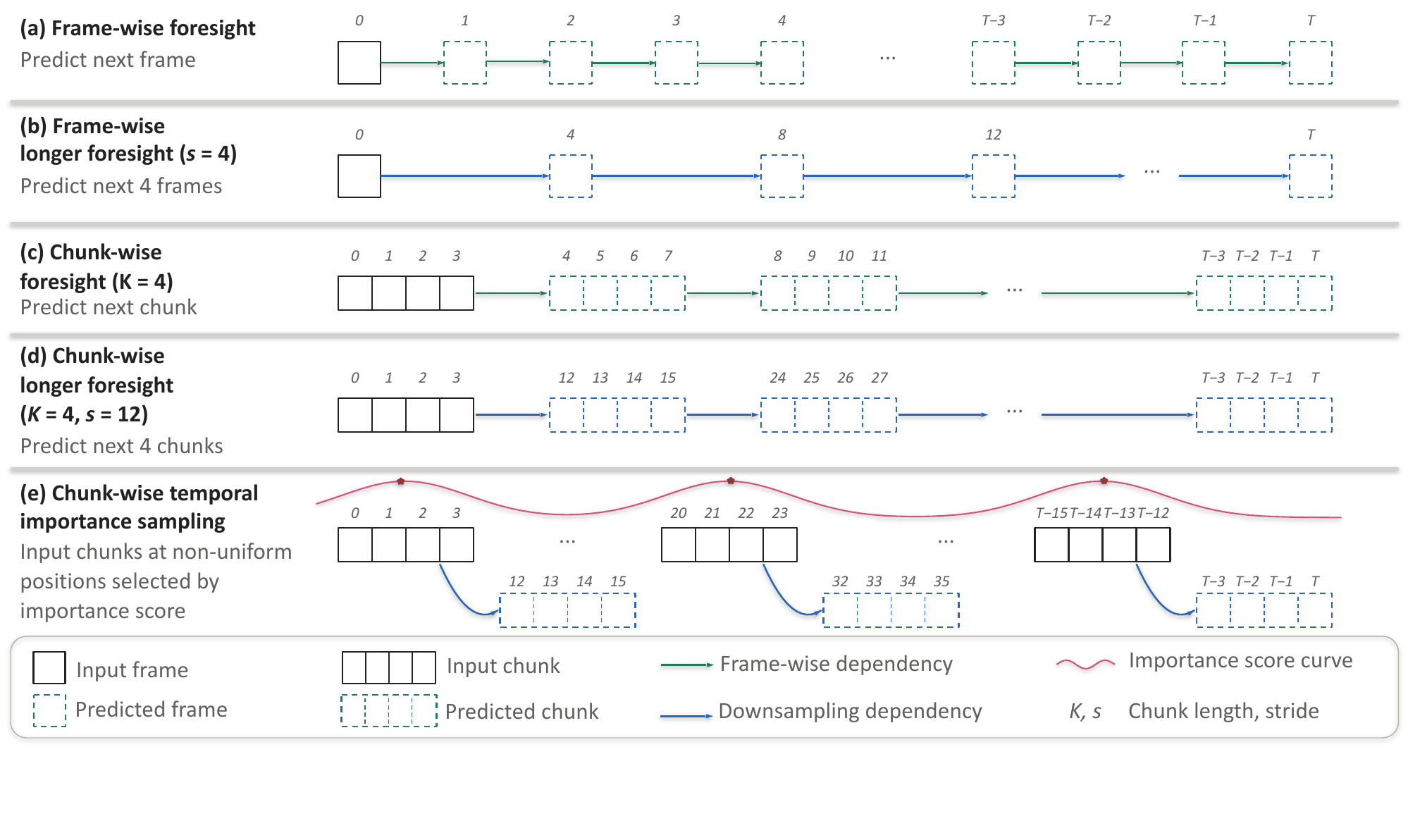}
    \vspace{-20pt}
    \caption{\small
    Prompt formulations for future frame prediction. 
    (a) Frame-wise foresight one frame at each step. 
    (b) Frame-wise longer foresight increased the temporal stride $s$. 
    (c) Chunk-wise foresight predicted a chunk of $K$ consecutive frames in parallel. 
    (d) Chunk-wise longer foresight combined chunk length $K$ with stride $s$ for longer-horizon prediction.
    (e) Chunk-wise temporal importance sampling.
    }
\label{fig:video_chunk}
\end{figure}

\paragraph{Chunk-Wise Prediction}
In \ldmname{} (LDM), we adopted a chunk-wise formulation in which the prompts were segmented into short, fixed-length units (typically of 1 second duration) that were predicted by the model in an auto-regressive manner. This design was motivated by both modeling considerations and computational efficiency. As illustrated in Fig~\ref{fig:video_chunk}~(a), frame-wise foresight provided only a weak learning signal at each step, since adjacent camera frames typically differed only marginally over short time intervals. A simple alternative (Fig~\ref{fig:video_chunk}~(b)) was to downsample the frame sequence and perform frame-wise longer foresight at a coarser temporal resolution. However, this could degrade performance because trajectory prediction relied on temporally coherent motion cues, which might be lost or aliased under aggressive temporal downsampling. In contrast, chunk-wise foresight, shown in Fig~\ref{fig:video_chunk}~(c), preserved short-term temporal structure within each chunk while requiring the model to predict a longer segment of future evolution at each auto-regressive step.

\paragraph{Curriculum Learning with Extended Foresight}
To further improve long-horizon prediction, we adopted a short-to-long curriculum learning strategy. Training started with short-horizon sequences in which adjacent chunks were temporally contiguous and separated by 1 second. We then extended the effective prediction horizon by increasing the temporal stride between neighboring chunks to 3 seconds, which was called Chunk-wise longer foresight as shown in Fig~\ref{fig:video_chunk}~(d). This larger temporal gap between chunks extended the prediction horizon while it did not increase the computational budgets. 

Under this temporally strided setting, the observation targets were predicted across the extended time gaps between chunks. In contrast, action prediction remained at the immediate next control step, since closed-loop control required temporally dense, unstrided trajectory outputs. This asymmetric design allowed the model to learn long-horizon visual evolution while preserving the temporal resolution required for ego-action prediction.

\paragraph{Temporal Importance Sampling} Chunk-wise longer foresight let LDM train on long horizons at tractable cost. A uniform sampler, however, wasted most of the budget on near-cruise segments and under-supervised the rare events that drove safety-critical decisions. We replaced it with an importance-weighted sampler as in Fig~\ref{fig:video_chunk}~(e).

For each candidate step $k$, we computed an importance score $w_k$ based on the longitudinal and lateral accelerations of the ego trajectory, denoted by $a_x$ and $a_y$, respectively. To capture safety-relevant dynamics at different temporal scales, we evaluated the trajectory over three temporal windows around the candidate step, denoted as $W_1^k$, $W_2^k$, and $W_3^k$. Within each window, we took the maximum value of a weighted acceleration magnitude, and then summed the resulting scores across windows:

\begin{equation}
w_k \;=\; \sum_{W \in \{W_1^k, W_2^k, W_3^k\}}
\max_{t \in W}\left(\lambda_x \lvert a_x(t) \rvert + \lambda_y \lvert a_y(t) \rvert\right).
\end{equation}
where $\lambda_x \geq 0$ and $\lambda_y \geq 0$ controlled the relative importance of longitudinal and lateral accelerations. The three temporal windows were designed to capture complementary maneuver phases. The near-future window $W_1^k$ emphasized imminent events such as braking, acceleration, or sudden swerve onsets. The mid-horizon window $W_2^k$ captured upcoming maneuver commitments, including turn-in and brake-in behavior. The recent-history window $W_3^k$ captured the aftermath of maneuvers that had just been executed. Together, these windows assigned higher sampling probability to steps that were temporally close to safety-relevant trajectory changes.

Given the per-step importance scores $\{w_k\}$, we sampled candidate steps according to a temperature-scaled distribution:
\begin{equation}
p_k \;=\; \frac{w_k^{1/\tau}}{\sum_{j} w_j^{1/\tau}},
\end{equation}
where $\tau > 0$ controlled the sharpness of the distribution. In practice, we additionally constrained the maximum temporal gap between consecutive selected steps. This prevented the sampled sequence from becoming overly sparse and ensured that future prediction remained well-conditioned on recent temporal context.

\paragraph{Semi-Causal Block Sparse Attention} 

\begin{figure}[h]
    \centering
    \includegraphics[width=0.8\textwidth]{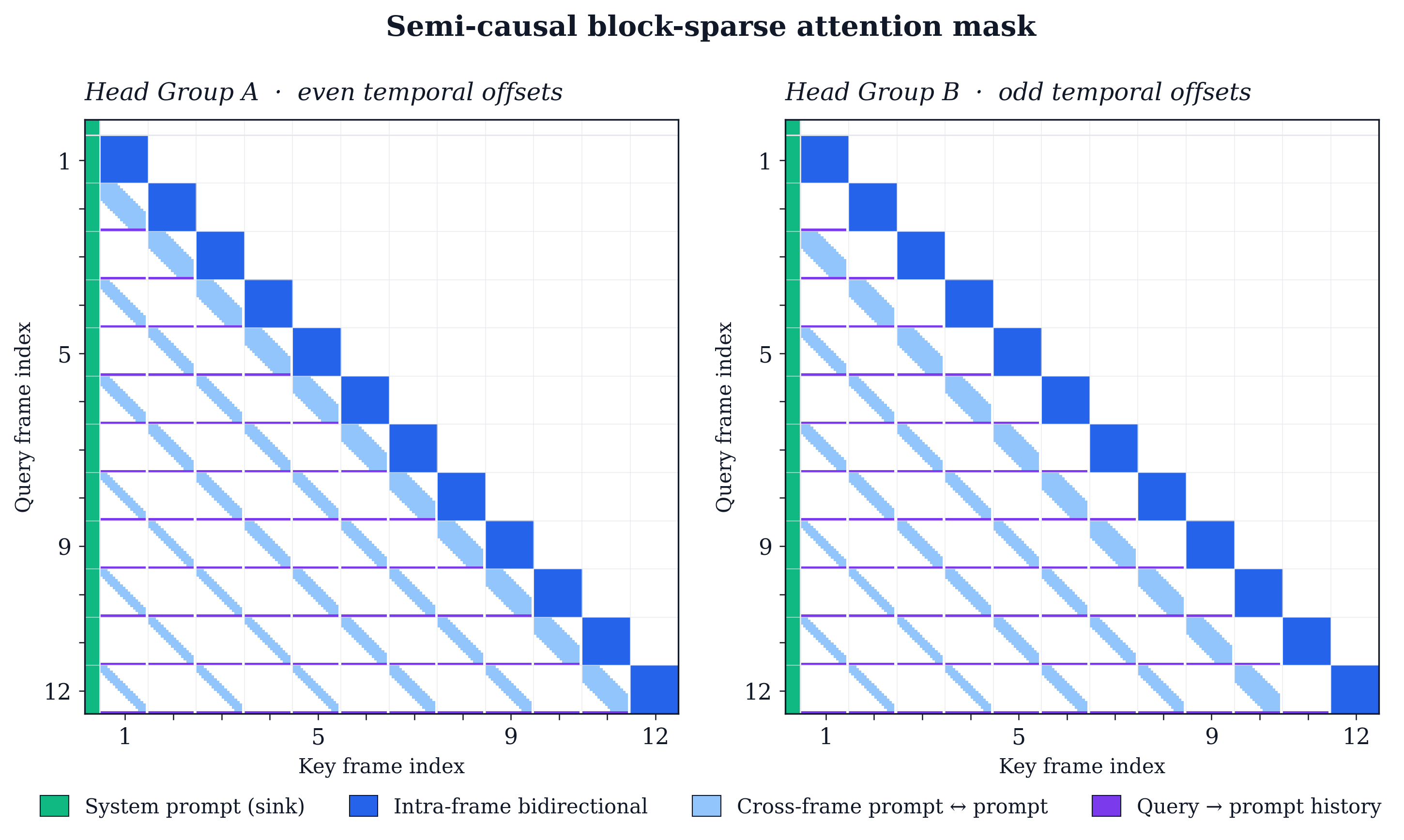}
    \caption{\small
    Semi-causal block-sparse attention mask for long-sequence training. 
    Each colored pixel denoted one token block, where attention was allowed. 
    The mask preserved bidirectional attention within each temporal chunk, allowed access to the global system prompt and previous prompt tokens, and prohibited attention between query tokens across different chunks. 
    The two panels showed complementary sparse patterns assigned to different attention-head groups.
    }
    \label{fig:bsa}
\end{figure}

To further speed up efficient long-horizon training, we adopted a semi-causal block-sparse attention mask implemented using block sparse attention~\cite{guo2024blocksparse}, inspired by the sparse attention patterns in~\cite{liradial}. The token sequence was first padded and partitioned into fixed-size blocks, and sparsity was imposed at the block level.

As illustrated in Fig~\ref{fig:bsa}, the attention mask was designed to preserve the temporal structure of the prediction task while reducing unnecessary long-range interactions. The system prompt was treated as a set of global sink tokens and was visible to all subsequent temporal chunks. Within each chunk, token blocks were connected with fully bidirectional self-attention, allowing the model to jointly reason over the text, observation, action/state, and query tokens associated with the same temporal window.

A temporally causal design was employed for interactions across different chunks. Because every chunk shared an identical internal layout, each prompt-side token, including $l_i$, $O_i$ and $A_i$, had a positional counterpart in every earlier chunk; cross-chunk attention was defined with respect to this counterpart, covering the same position and a surrounding spatial neighborhood whose width shrank with temporal distance; this design allowed both temporal tracking and spatial reasoning while keeping total compute manageable. 
Query tokens $Q_i$ retained full access to all preceding prompt-side tokens, enabling future predictions to condition on the available observation and action history. However, attention from $Q_i$ to previous query tokens $Q_{1:i-1}$ was masked to prevent predictions from directly depending on earlier predicted targets.

To further reduce computation, we partitioned attention heads into two disjoint groups based on the parity of the timestep difference between query and key blocks. Each group attended to a complementary subset of temporal offsets, which reduced the number of active block connections while maintaining coverage over the temporal context. As a result, the number of attended blocks grew approximately linearly with sequence length, rather than quadratically.

\subsubsection{Training Objectives}\label{subsubsec:training-objectives}
During training, we optimized the model to predict future observations and ego actions under a teacher-forcing formulation. Specifically, ground-truth observations and actions were encoded and inserted into the multi-modal prompt for previous temporal chunks, while the model was trained to predict the targets at future prediction steps. The training loss was applied to both the predicted observation tokens $\hat{\mathbf{o}}_i^v$ and the decoded ego actions $\hat{\mathbf{a}}_i$.

\begin{equation}
L_{\mathrm{cam}} =
\frac{1}{H V}\sum_{i=1}^{H}\sum_{v=1}^{V}\left\|\hat{\mathbf{o}}_i^v - g(\mathbf{I}_i^v)\right\|_2 ,
\end{equation}

\begin{equation}
L_{\mathrm{act}} =\frac{1}{H}\sum_{i=1}^{H}\left\|\hat{\mathbf{a}}_i - \mathbf{a}_i\right\|_1 .
\end{equation}
where $H$ was the total horizon used in training, $V$ was the number of camera views. $g(\cdot)$ was the frozen ViT from the base vision language model. 

We additionally incorporated an auxiliary bird's-eye-view (BEV) prediction loss during training to encourage the model to learn geometrically meaningful scene representations. The predicted BEV map $\hat{\mathbf{b}}_i$ was decoded from a subset of the query tokens in $Q_i$ and supervised by the corresponding ground-truth BEV target $\mathbf{b}_i$:
\begin{equation}
L_{\mathrm{bev}} = \frac{1}{H} \sum_{i=1}^{H}\left\|\hat{\mathbf{b}}_i - \mathbf{b}_i\right\|_2 .
\end{equation}
The final training objective was a weighted sum of the action, camera, and BEV losses:
\begin{equation}\label{eq:total_loss_ldm}
L_{\mathrm{total}} = L_{\mathrm{act}} + \alpha L_{\mathrm{cam}} + \beta L_{\mathrm{bev}},
\end{equation}
where $\alpha$ and $\beta$ controlled the relative weights of the visual prediction and BEV prediction losses, respectively.

\subsection{\vrname}
The Vision Renderer was the terminal module of \modelname's auto-regressive pipeline: it converted the LDM's predicted camera latent tokens into high-fidelity multi-view future frames that could be fed back into the vision encoder to close the inference loop. Camera latent tokens were optimized for
  auto-regressive reasoning and therefore compressed away much of the high-frequency appearance detail
  required for photorealistic closed-loop feedback; decoding them directly in the vision decoder would
  accumulate reconstruction error over long rollouts and degrade downstream perception. Unlike prior
  generative world models that coupled action-conditioned dynamics with pixel synthesis in a single network,
   we intentionally separated the two roles --- imagination and control lived in the LDM, while the renderer
   acted as a conditional refiner that, conditioned on a short window of history images and the
  LDM-predicted camera tokens, restored the high-frequency visual detail lost during latent compression.
\subsubsection{Design Rationale}
  A central design decision was that the renderer was not conditioned on the LDM's action tokens. The camera
   tokens already encoded the ego-vehicle transformation implicitly, together with surrounding scene
  geometry and dynamic agents; in our setting we found that the camera token was the only LDM-produced conditioning signal required, with multi-view history continuing to enter through the standard I2V pathway and no additional action conditioning needed. We adopted this restriction to avoid the risk that, if
   both signals were exposed, the renderer might exploit action tokens as a low-entropy shortcut and ignore
  the camera tokens, causing the LDM and renderer to predict futures independently and defeating the purpose
   of the closed loop. As a consistency check, we compared the renderer's frames against the pixel-space
  decode of the same camera tokens produced by the Camera Latent Decoder (Section~\ref{sec:cld}); their agreement was
   consistent with the camera token acting as the effective information bottleneck through which the LDM
  controlled the rendered future.

\subsubsection{Model Architecture}

\paragraph{Backbone} The Vision Renderer was built upon X-World, a Diffusion Transformer (DiT)~\cite{peebles2023scalable} video generator that coupled a 3D causal VAE with a latent denoiser trained under a rectified-flow objective~\cite{lipman2023flow}. The encoder and decoder were based on the 3D causal VAE from the WAN2.2 release of WAN~\cite{wan2025wan}, which compressed input videos along both spatial and temporal dimensions into a compact latent space for efficient denoising. Rectified flow learned a time-dependent velocity field $v_\theta(\mathbf{y}_t, t, \mathbf{c})$ that mapped a Gaussian prior to the data distribution along a straight interpolation path
\begin{equation}
\mathbf{y}_t = (1-t)\mathbf{y}_0 + t\,\mathbf{y}_1,\quad \mathbf{y}_0 \sim p_{\mathrm{data}}(\mathbf{y}\mid \mathbf{c}),\ \mathbf{y}_1 \sim \mathcal{N}(\mathbf{0},\mathbf{I}),\ t\sim\mathcal{U}(0,1),
\label{eq:rf_interp}
\end{equation}
by minimizing
\begin{equation}
\mathcal{L}_{\mathrm{velocity}}(\theta)=\mathbb{E}_{\mathbf{y}_0,\mathbf{y}_1,t,\mathbf{c}}\left[\left\|v_\theta(\mathbf{y}_t,t,\mathbf{c})-(\mathbf{y}_1-\mathbf{y}_0)\right\|_2^2\right],
\label{eq:rf_loss}
\end{equation}
where $\mathbf{c}$ denoted the conditioning inputs passed to the renderer. Relative to standard denoising diffusion, rectified flow yielded straighter probability paths, enabling high-quality samples with fewer function evaluations.

\paragraph{Cross-view Attention} To enforce geometric consistency across synchronized surround-view cameras, we adopted the view-temporal self-attention design of X-World: self-attention was applied alternately along the temporal and cross-view axes over latent tokens from multiple cameras, encouraging coherent geometry, object identity, and motion across viewpoints while preserving temporal smoothness.

\paragraph{Conditioning} The renderer departed from X-World in what it conditioned on. X-World was a full generative simulator and therefore accepted vectorized conditions describing dynamic agents (bounding boxes, categories) and static road elements (lane lines, boundaries) alongside text prompts, actions, and camera parameters. In \modelname, these scene descriptors were already encoded by the LDM into the camera token stream: the tokens carried both the ego-vehicle pose and the predicted placement and behavior of surrounding agents. Consequently, in its final configuration the renderer retained only a single cross-attention branch
   that injected the LDM's camera tokens at each DiT block, with X-World's action, dynamic-agent,
  static-element, and text-conditioning branches all removed. The action-conditioning branch was
  retained temporarily during fine-tuning so that the renderer could be adapted under ground-truth
  ego actions before being spliced onto the LDM's camera-token stream
  (Section~\ref{subsubsec:vr-training}). Multi-view history frames were provided through the standard
   image-to-video (I2V) input pathway of WAN.

\paragraph{Temporal Alignment} Most pre-trained video generation models operated at 12~Hz, whereas \modelname's LDM emitted one camera token per rendered frame at 4~Hz, matching the output rate of the action policy head. We adapted the renderer to 4~Hz during fine-tuning (Section~\ref{subsubsec:vr-training}) so that each rendered frame was tightly anchored one-to-one to a corresponding LDM camera-token prediction.

\paragraph{Rollout Drift Mitigation} Under closed-loop deployment, each
rendered frame became part of the conditioning history for the next step,
exposing the renderer to its own samples rather than ground-truth frames.
This distributional gap accumulated over the rollout and caused predicted
frames to drift off the data manifold. Recent works such as
CausVid~\cite{yin2025slow} and Self-Forcing~\cite{huang2026self}
addressed this via distribution matching distillation (DMD). For
\modelname, we instead adopted two lighter-weight modifications: a latent
sink that anchored a stable reference context across rollout steps, and
latent augmentation on the current-step latent during training, following
Helios~\cite{yuan2026helios}, which exposed the renderer to perturbed
conditioning resembling its inference-time distribution.

\subsection{Pipeline}
\subsubsection{Training}
\modelname{} coupled a \ldmname{} (LDM), which jointly modeled actions and
high-level visual semantics, with a diffusion-based vision renderer that
produced photorealistic multi-view frames. Training proceeded in three
stages. In stage~I, the LDM was trained alone under teacher forcing to
emit future actions together with compact camera tokens that summarized
the mean of plausible future appearances. In stage~II, the vision renderer
was independently adapted to synthesize photorealistic multi-view imagery
conditioned on ground-truth future ego actions. In stage~III, the LDM was
\emph{frozen} and only the renderer was further fine-tuned, with its
action-conditioning branch swapped for the LDM's predicted future camera
tokens, so that the renderer learned to produce frames consistent with the
token distribution it would actually consume under the closed-loop
auto-regressive rollout used at inference. This decoupled-then-aligned
schedule kept LDM and renderer training tractable in stages~I and~II while
letting stage~III close the gap between training-time teacher forcing and
inference-time rollout, without disturbing the LDM's learned action and
semantic predictions.

\paragraph{Stage~I: LDM Pretraining}
We trained the \ldmname{} (LDM) under teacher forcing to jointly predict future ego actions, multi-view camera tokens, and an auxiliary BEV latent, using the chunk-wise multi-modal prompt and the combined loss $L_{\mathrm{total}} = L_{\mathrm{act}} + \alpha L_{\mathrm{cam}} + \beta L_{\mathrm{bev}}$. Three components made this stage tractable on long horizons: a semi-causal block-sparse attention mask, with heads partitioned by parity so block density grew linearly with sequence length; a short-to-long curriculum that progressively increased the inter-chunk stride from 1~s to 3~s, shifting supervision from near-frame extrapolation toward longer-horizon dynamics; and an importance-weighted temporal sampler that scored candidate AR steps by peak weighted longitudinal/lateral acceleration, concentrating supervision on safety-critical futures. Because each predicted camera token was regressed against many plausible futures under an L2 objective, it converged to a low entropy summary of the conditional future --- enough for action grounding and scene structure, but visually blurry when decoded directly.

\paragraph{Stage~II: Renderer Pretraining}
In parallel, the renderer was adapted to synthesize photorealistic multi-view frames conditioned on ground-truth ego actions, rather than on camera tokens predicted by the LDM. We initialized from X-World, a DiT video generator that already provided cross-view temporal attention, and adapted it in two steps. \emph{(i) Temporal alignment.} We fine-tuned the renderer from WAN's native 12~Hz to the LDM's 4~Hz emission rate, so each generated frame was anchored one-to-one to a single LDM-predicted token in stage~III. \emph{(ii) Rollout-drift mitigation.} We attached a latent sink that anchored a stable reference context across rollout steps, and applied latent augmentation to the current-step latent during training so the renderer saw perturbed conditioning resembling its inference-time distribution.

\paragraph{Stage~III: Renderer Alignment}
In the third stage, the LDM was frozen and the vision renderer was
further finetuned so that rendered frames stayed consistent with the camera
tokens the LDM actually produced under the closed-loop rollout used
at inference. The renderer now conditioned on camera tokens
\emph{predicted} by the LDM rather than ground-truth actions, and
was supervised solely by the rectified-flow velocity-matching objective
$\mathcal{L}_{\mathrm{velocity}}$ in Eq.~(\ref{eq:rf_loss}). The stage-I
action and BEV losses on the LDM were dropped: the LDM no
longer received gradients, and spatial structure was anchored through the
renderer's pixel-space supervision. Finetuning at the 4~Hz cadence
established in stage~II closed the train/inference gap left by teacher
forcing and yielded the tightly interleaved prediction-rendering loop
deployed at inference time.

\subsubsection{Inference}
At inference time, \modelname{} performed an auto-regressive rollout that strictly interleaved prediction and rendering at the 4~Hz cadence established in stage~III. Each step began with a text prompt and a short window of multi-view history frames. The LDM consumed this context using the same chunk-wise prompt format as in training and emitted, in a single forward pass, the next ego action and one second of multi-view latent tokens — four frames per camera at 4~Hz, all seven camera views per frame, each camera-frame represented by several tokens — summarizing the predicted future appearance from each viewpoint. Decoding the camera token directly through the Camera Latent Decoder yielded a blurry image, because the token represented the mean of plausible futures rather than any specific realization. The vision renderer then conditioned on this token together with the multi-view history and ran a small number of rectified-flow sampling steps to draw one concrete future from the implied distribution, producing high-fidelity, view-consistent surround imagery. Because the renderer was conditioned only on the camera token --- not on the predicted action --- the rendered future was fully determined by the LDM's latent imagination, which also let us audit alignment by comparing the renderer's frames against the Camera Latent Decoder's decode of the same token. The rendered frames and the new action were appended to the rolling context and fed back into the LDM for the next step. Iterating this loop produced arbitrarily long closed-loop rollouts of paired action trajectories and photorealistic surround-view video.

\section{Experiments}
This section presented the implementation details of both LDM and \vrname{}. Comprehensive experiments and ablation studies were conducted on our in-house dataset.
\subsection{\ldmname}
\subsubsection{Trajectory Results}

This section evaluated \modelname{} along three axes. We first examined the effect of training with longer chunk-wise horizons. We then ablated three refinements of long-horizon training: curriculum learning (CL), curriculum learning with extended foresight (CLEF) and temporal importance sampling (TIS). Finally, we reported the combined system at production scale, both quantitatively and through two qualitative scenarios.

We reported two complementary sets of metrics. ADE and FDE (in meters) measured the displacement between predicted and ground-truth ego trajectories. The CCES suite measured driving quality across four axes: \emph{Compliance} (traffic-rule adherence, e.g., lane centering, line crossings, red-light running), \emph{Comfort} (ride quality, e.g., steering instability, motion sickness), \emph{Efficiency} (progress toward destination, e.g., under-acceleration, failure to overtake slow vehicles), and \emph{Safety} (avoidance of dangerous events, e.g., collisions, time-to-collision violations, hitting road boundaries). Each underlying metric was a per-frame fail-rate proportion. To put metrics of very different magnitudes on a common scale, we took the first row of Table~\ref{tab:ar_horizon} as the shared reference and reported every other entry across all three tables as its ratio to that reference; each CCES category cell was then the unweighted mean of these ratios across the curated metrics in that category, and Total was the sum of the four category means. By construction the reference row therefore had $1$ in every category cell and $4$ in the Total column, and lower was better everywhere. Collision rate was additionally reported as a standalone safety headline in its raw percentage fail-rate form. To ensure a fair comparison, all models were evaluated using only a single forward step at inference time, isolating the effect of training-time supervision on single-step inference quality. 

\paragraph{Effect of Long-Horizon Training.}
Table~\ref{tab:ar_horizon} evaluated the effect of scaling the training-time horizon $H \in \{1, 6, 21\}$, where each step predicted one chunk of one second. This covered no long-horizon supervision ($H{=}1$, single-chunk training), a moderate horizon ($H{=}6$), and our full long-horizon configuration ($H{=}21$). All three rows shared architecture, data, hardware (128 GPUs), and number of training steps; only $H$ varied. ADE, FDE, and the standalone collision rate all improved monotonically with $H$, dropping by roughly 2--7\% from $H{=}1$ to $H{=}21$. The Safety ($1.0000 \rightarrow 0.9481$) and Compliance ($1.0000 \rightarrow 0.9533$) categories, the axes most directly aligned with long-horizon world-causality supervision, also improved monotonically. Comfort and Efficiency saw small regressions at $H{=}21$, revealing the optimization difficulty of naively scaling $H$; these partially offset the Safety/Compliance gains in the summed Total, which dipped from $4.0000$ at $H{=}1$ to $3.9396$ at $H{=}6$ but regressed marginally to $3.9524$ at $H{=}21$. The overall direction confirmed our central claim that long-horizon chunk-wise supervision benefited the safety- and compliance-related axes, and the small Total regression at $H{=}21$ motivated the curriculum-learning and importance-sampling designs ablated next.

\begin{table}[h]
\centering
\caption{Effect of training-time horizon $H$. All models shared architecture, data, hardware (128 GPUs), and number of training steps; only $H$ varied. ADE/FDE in meters; Coll. is the percentage collision fail rate (\%); Compl., Comfort, Eff., and Safety are each the unweighted mean of per-metric ratios to $H{=}1$ (lower $=$ better); Total sums the four category means and equals 4 at $H{=}1$ by construction. Lower is better throughout.}
\label{tab:ar_horizon}
\small
\setlength{\tabcolsep}{3pt}
\begin{tabular}{l|cc cc|c|c c c c|c}
\toprule
\multirow{2}{*}{Method} & \multicolumn{2}{c}{ADE $\downarrow$} & \multicolumn{2}{c|}{FDE $\downarrow$} & \multirow{2}{*}{Coll.\ $\downarrow$} & \multirow{2}{*}{Compl.\ $\downarrow$} & \multirow{2}{*}{Comfort $\downarrow$} & \multirow{2}{*}{Eff.\ $\downarrow$} & \multirow{2}{*}{Safety $\downarrow$} & \multirow{2}{*}{\textbf{Total $\downarrow$}} \\
& Lat. & Long. & Lat. & Long. & & & & & & \\
\midrule
$H{=}1$  & 0.1923 & 1.2409 & 0.4881 & 3.1935 & 0.263 & 1.0000 & 1.0000 & 1.0000 & 1.0000 & 4.0000 \\
$H{=}6$  & 0.1864 & 1.2196 & 0.4691 & 3.1178 & 0.262 & 0.9756 & 0.9880 & 0.9833 & 0.9927 & 3.9396 \\
$H{=}21$ & 0.1810 & 1.2110 & 0.4571 & 3.0988 & 0.245 & 0.9533 & 1.0416 & 1.0094 & 0.9481 & 3.9524 \\
\bottomrule
\end{tabular}
\end{table}

\paragraph{Ablating CL, CLEF and TIS.}
Table~\ref{tab:subsample_ablation} ablated how the $H{=}21$ horizon should be supervised. All four rows were initialized from the $H{=}6$ checkpoint and additionally trained on 128 GPUs; only the second-stage strategy differed. Row~1 (Cont.\ $H{=}6$) continued training at $H{=}6$. Row~2 ($+\,H{=}21$, CL) applied basic curriculum learning (short-to-long $H$ schedule at fixed 1\,s inter-chunk stride). Row~3 ($+\,H{=}21$, CLEF) applied our curriculum learning with extended foresight, progressively widening the inter-chunk stride from 1\,s to 3\,s. Row~4 ($+\,H{=}21$, TIS) added our temporal importance sampling on top of CLEF, concentrating supervision on safety-critical chunks identified by peak weighted lateral/longitudinal acceleration. Continuing $H{=}6$ (Row~1) reduced ADE/FDE relative to the $H{=}6$ checkpoint while collision rate plateaued ($0.262 \rightarrow 0.270$), but did not match Rows~3 and 4, ruling out additional training time as the sole explanation. Row~2 (CL) further reduced collision rate ($0.270 \rightarrow 0.238$) and Total CCES ($3.9523 \rightarrow 3.8745$). CLEF (Row~3) delivered the strongest ADE/FDE gains (lowest lateral ADE/FDE), indicating that progressively extending the foresight horizon stabilized trajectory prediction, though Safety regressed slightly versus CL ($0.9310 \rightarrow 0.9387$), motivating the safety-targeted sampler in Row~4. TIS (Row~4) further reduced the collision rate from $0.230$ to $0.216$, a $6.1\%$ relative reduction, and achieved the lowest Total CCES ($3.8447$); the Safety ratio improved to $0.9264$, the lowest among the four rows, consistent with the design intent of biasing supervision toward safety-critical chunks.

\begin{table}[h]
\centering
\caption{Ablating CL, CLEF, and TIS. All four rows were initialized from the $H{=}6$ checkpoint and additionally trained on 128 GPUs. ADE/FDE in meters; Coll. is the percentage collision fail rate (\%); Compl., Comfort, Eff., and Safety are unitless ratios as in Table~\ref{tab:ar_horizon}. Lower is better. \emph{Cont.}: continue training at $H{=}6$; \emph{CL}: basic curriculum learning; \emph{CLEF}: curriculum learning with extended foresight; \emph{TIS}: temporal importance sampling.}
\label{tab:subsample_ablation}
\small
\setlength{\tabcolsep}{3pt}
\begin{tabular}{l|cc cc|c|c c c c|c}
\toprule
\multirow{2}{*}{Method} & \multicolumn{2}{c}{ADE $\downarrow$} & \multicolumn{2}{c|}{FDE $\downarrow$} & \multirow{2}{*}{Coll.\ $\downarrow$} & \multirow{2}{*}{Compl.\ $\downarrow$} & \multirow{2}{*}{Comfort $\downarrow$} & \multirow{2}{*}{Eff.\ $\downarrow$} & \multirow{2}{*}{Safety $\downarrow$} & \multirow{2}{*}{\textbf{Total $\downarrow$}} \\
& Lat. & Long. & Lat. & Long. & & & & & & \\
\midrule
Cont.\ $H{=}6$  & 0.1741 & 1.1807 & 0.4344 & 3.0087 & 0.270 & 0.9302 & 1.0515 & 0.9980 & 0.9726 & 3.9523 \\
+$H{=}21$, CL   & 0.1718 & 1.1671 & 0.4277 & 2.9856 & 0.238 & 0.9326 & 1.0106 & 1.0003 & 0.9310 & 3.8745 \\
+$H{=}21$, CLEF & 0.1692 & 1.1571 & 0.4181 & 2.9421 & 0.230 & 0.9320 & 1.0076 & 0.9951 & 0.9387 & 3.8734 \\
+$H{=}21$, TIS  & 0.1696 & 1.1578 & 0.4195 & 2.9413 & 0.216 & 0.9187 & 1.0043 & 0.9953 & 0.9264 & 3.8447 \\
\bottomrule
\end{tabular}
\end{table}

\paragraph{Production-Scale Comparison.}
Table~\ref{tab:headline} reported the headline comparison at production scale, combining long-horizon training ($H{=}21$) with CLEF and TIS, trained on 1024 GPUs against the baseline trained at the same scale. The full \modelname{} model reduced lateral/longitudinal ADE by $6.4\%/3.6\%$ ($0.1675 \rightarrow 0.1567$ lat; $1.1387 \rightarrow 1.0982$ long) and lateral/longitudinal FDE by $8.8\%/4.1\%$ ($0.4153 \rightarrow 0.3789$ lat; $2.9117 \rightarrow 2.7924$ long). Collision rate dropped from $0.228\%$ to $0.191\%$, a $16.2\%$ relative reduction. All four CCES categories improved: Safety by $9.1\%$, Compliance by $8.2\%$, Comfort by $1.0\%$, and Efficiency by $0.4\%$. The aggregate Total CCES improved from $3.8296$ to $3.6535$, a $4.6\%$ relative reduction, with the gains concentrated in Safety and Compliance, consistent with the central claim that long-horizon world-causality supervision improved safety-critical decision quality.

\begin{table}[h]
\centering
\caption{Production-scale comparison against the baseline, both trained on 1024 GPUs. \modelname{} (Ours) used $H{=}21$ with CLEF and TIS. ADE/FDE in meters; Coll. is the percentage collision fail rate (\%); Compl., Comfort, Eff., and Safety are unitless ratios as in Table~\ref{tab:ar_horizon}. Lower is better.}
\label{tab:headline}
\small
\setlength{\tabcolsep}{3pt}
\begin{tabular}{l|cc cc|c|c c c c|c}
\toprule
\multirow{2}{*}{Method} & \multicolumn{2}{c}{ADE $\downarrow$} & \multicolumn{2}{c|}{FDE $\downarrow$} & \multirow{2}{*}{Coll.\ $\downarrow$} & \multirow{2}{*}{Compl.\ $\downarrow$} & \multirow{2}{*}{Comfort $\downarrow$} & \multirow{2}{*}{Eff.\ $\downarrow$} & \multirow{2}{*}{Safety $\downarrow$} & \multirow{2}{*}{\textbf{Total $\downarrow$}} \\
& Lat. & Long. & Lat. & Long. & & & & & & \\
\midrule
Baseline & 0.1675 & 1.1387 & 0.4153 & 2.9117 & 0.228 & 0.9483 & 0.9505 & 0.9867 & 0.9441 & 3.8296 \\
X-Foresight     & 0.1567 & 1.0982 & 0.3789 & 2.7924 & 0.191 & 0.8708 & 0.9413 & 0.9831 & 0.8583 & 3.6535 \\
\bottomrule
\end{tabular}
\end{table}

\begin{figure*}[t]
    \centering
    \includegraphics[width=\linewidth]{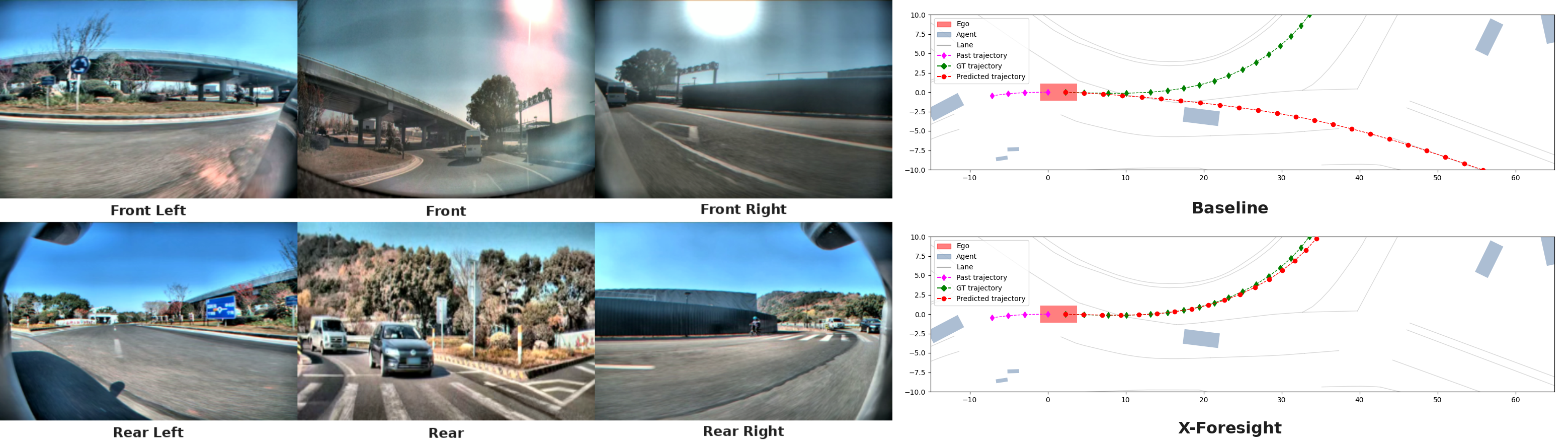}\\
    \vspace{-7pt}
    \tikz{\draw[dashed, gray] (0,0) -- (\linewidth,0);}\\[3pt]
    \includegraphics[width=\linewidth]{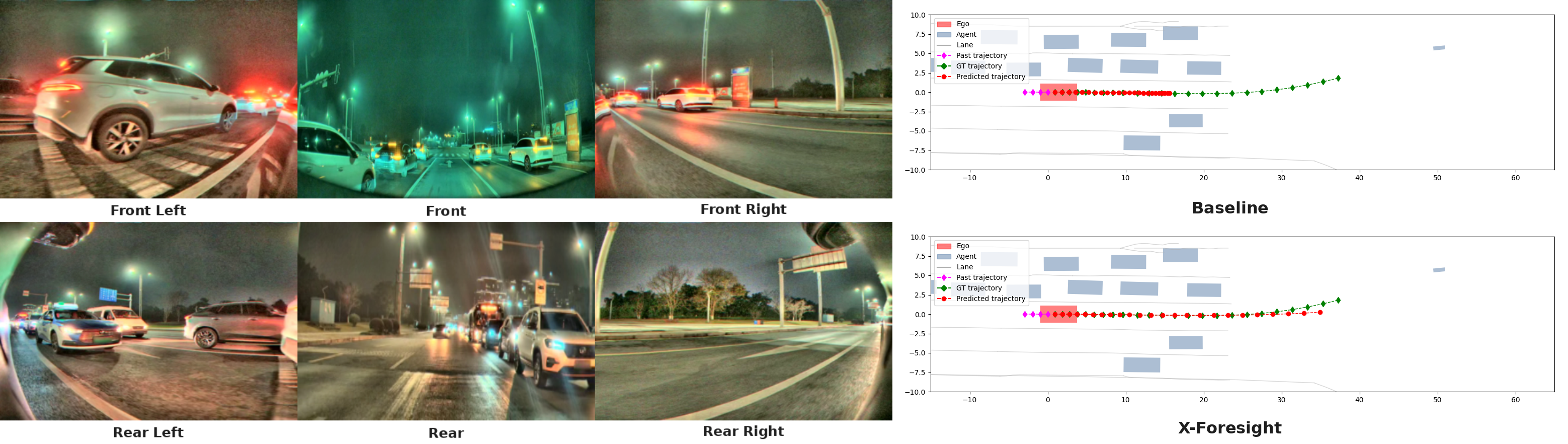}
    \caption{Qualitative comparisons. The baseline failed on events lying ahead in space (top, a far-exit instruction in a multi-exit roundabout) or in time (bottom, a red light that turned green before arrival); \modelname{} tracked the ground truth in both.}
    \label{fig:qualitative}
\end{figure*}

The two scenarios in Fig~\ref{fig:qualitative} illustrated how this aggregate gain manifested qualitatively, both turning on correct action toward events that lay ahead in space or in time, the central capability the baseline lacked by construction. In the top panel, the ego entered a multi-exit roundabout under a navigation instruction that directed it to take a later exit rather than the first one. The baseline committed to the closest visible exit and emitted a trajectory that veered off the lane, diverging substantially from the ground truth. \modelname{} held the roundabout lane and tracked the ground truth all the way to the intended exit, anchoring its action on the long-horizon navigation target rather than on the most salient local cue. In the bottom panel, the ego approached a nighttime intersection while the traffic light was still red at the current frame; the underlying log confirmed that the light would turn green before the ego reached the stop line, so the ground-truth trajectory continued through the intersection without braking. The baseline treated the present red signal as an immediate stop and truncated its prediction well before the stop line. \modelname{} anticipated the signal transition through its predictive rollout and produced a continuous trajectory matching the ground truth.

\subsubsection{Acceleration}

We optimized the Block Sparse Attention (BSA) kernel for our target hardware and integrated it as a drop-in replacement for FlashAttention-2 (FA2). Table~\ref{tab:bsa_acc} reported the wall-clock time per training step before and after this change, with the FA2 baseline using a standard causal mask. Per-step time was measured under identical model and training configurations, with the attention implementation and associated mask being the only changes. Pairing BSA with our tailored sparsity pattern reduced per-step time from 24.50\,s to 15.40\,s, a 1.59$\times$ speedup, confirming that attention was the major bottleneck and that the savings from our sparsity design translated directly into end-to-end training throughput.

\begin{table}[t]
\centering
\caption{Training throughput comparison over long-horizon sequences.}
\label{tab:bsa_acc}
\begin{tabular}{lcc}
\toprule
Attention implementation & Per-step time (s) $\downarrow$ & Speedup $\uparrow$ \\
\midrule
FlashAttention-2~\cite{dao2024flashattention}      & 24.50 & 1.00$\times$ \\
BSA w/ mask (ours) & \textbf{15.40} & \textbf{1.59$\times$} \\
\bottomrule
\end{tabular}
\end{table}

\subsubsection{Camera Latent Decoder}
\label{sec:cld}

In long-horizon rollouts, future observations were generated recursively from the model’s own decoded latent, making the structure and consistency of the latent space critical to stable future prediction and downstream planning performance. However, our vision renderer was diffusion-based, where rendered images were influenced by strong generative priors and sampling stochasticity rather than being deterministic projections of the latent representation. Consequently, rendered outputs could not reliably reflect the underlying latent space.

To address this issue, we introduced a lightweight Camera Latent Decoder to visualize the predicted latent space, following prior works\cite{Gao2025OneLI}\cite{shi2025}. Unlike diffusion-based rendering, our latent decoder provided a more faithful and stable projection of the latent representation, enabling direct inspection of spatial structure, temporal consistency, and multi-view alignment. This capability was essential for diagnosing latent drift, understanding error accumulation in long-horizon rollouts, and ensuring that the learned latent space satisfied the structural assumptions required by the downstream vision renderer.

To enable direct inspection of the predicted latent space, we trained a lightweight latent decoder that reconstructed image sequences from the model’s spatio-temporal latent representations. Given latent representations extracted from the world model, the decoder progressively upsampled along both temporal and spatial dimensions to reconstruct high-resolution image frames from the compressed latent representation. The decoder adopted a lightweight convolutional architecture with progressive spatio-temporal upsampling based on causal 3D ResNet blocks. A lightweight spatial self-attention module was introduced at the lowest-resolution stage to enable global feature interaction, allowing efficient reconstruction of high-resolution image sequences from the predicted latent representations.

\subsection{\vrname}
\subsubsection{Training Process}\label{subsubsec:vr-training}

We initialized the Vision Renderer from X-World weights,
retaining all transferable parameters of the DiT backbone and the 3D causal
VAE. The newly introduced camera-token cross-attention branch was randomly
initialized, while the action, dynamic-agent, static-element, and
text-conditioning branches inherited from X-World were discarded in stage~III.
Training data consisted of synchronized multi-view driving clips paired with
LDM-produced camera tokens for the same intervals, so that the renderer learned
to denoise future frames under the exact token distribution it would consume at
inference.

To bridge the gap between X-World's pre-training regime and the closed-loop
deployment setting of \modelname, we followed a decoupled-then-aligned
schedule spanning the two renderer training stages: in stage~II the renderer
was adapted independently of the LDM under ground-truth action conditioning,
while in stage~III the LDM was frozen and only the renderer was fine-tuned
to consume the LDM's predicted camera tokens.

\paragraph{Stage~II: Renderer Pretraining.} We first fine-tuned the diffusion
renderer on the multi-view driving corpus to adapt it from X-World's 12~Hz
pre-training cadence to the 4~Hz cadence emitted by \modelname's LDM. We
optimized the rectified-flow objective on subsampled video streams with the VAE
temporal stride adjusted accordingly. The LDM was frozen throughout this stage.
The renderer retained X-World's original action-conditioning branch (consuming
ground-truth ego actions rather than LDM-predicted camera tokens); the
camera-token cross-attention branch that would replace it in stage~III was not
yet active. We used the Muon optimizer with a constant learning rate of
$8\times10^{-5}$, batch size $1$ per device, and $128$ GPUs.

\paragraph{Stage~III: Renderer Alignment.} In the third stage, we spliced the
renderer onto the LDM, swapped the conditioning source, and froze the LDM so
that only the renderer received gradients. The action-conditioning branch
used during stage~II, together with X-World's dynamic-agent, static-element,
and text-conditioning branches, were all removed; the renderer was exposed
exclusively to (i) multi-view history latents and (ii) LDM-produced camera
tokens through the newly activated cross-attention branch, committing the
DiT to the LDM's latent imagination as its sole control signal. The renderer was then finetuned under the rectified-flow
velocity-matching objective $\mathcal{L}_{\mathrm{velocity}}$ in
Eq.~(\ref{eq:rf_loss}), now conditioned on the LDM's predicted camera
tokens. To fit the model into device memory, we
applied activation checkpointing on every DiT block and sharded the
renderer's parameters, gradients, and optimizer states with FSDP; the LDM
was kept in inference mode and contributed only forward activations. We
optimized with Muon under a one-cycle cosine learning-rate schedule, tuned
for stable convergence of the diffusion velocity field. From stage~III onward we monitored the controllability of the renderer by periodically
comparing its samples against the Camera Latent Decoder's pixel decode of
the same camera tokens, as described in Section~\ref{sec:cld}; the renderer's samples consistently aligned with the Camera Latent Decoder's pixel decode of the same camera tokens, confirming that the rendered frames faithfully tracked the LDM's camera-token condition.

\subsubsection{Results}
\begin{figure}[ht]
  \centering
  \includegraphics[width=\linewidth]{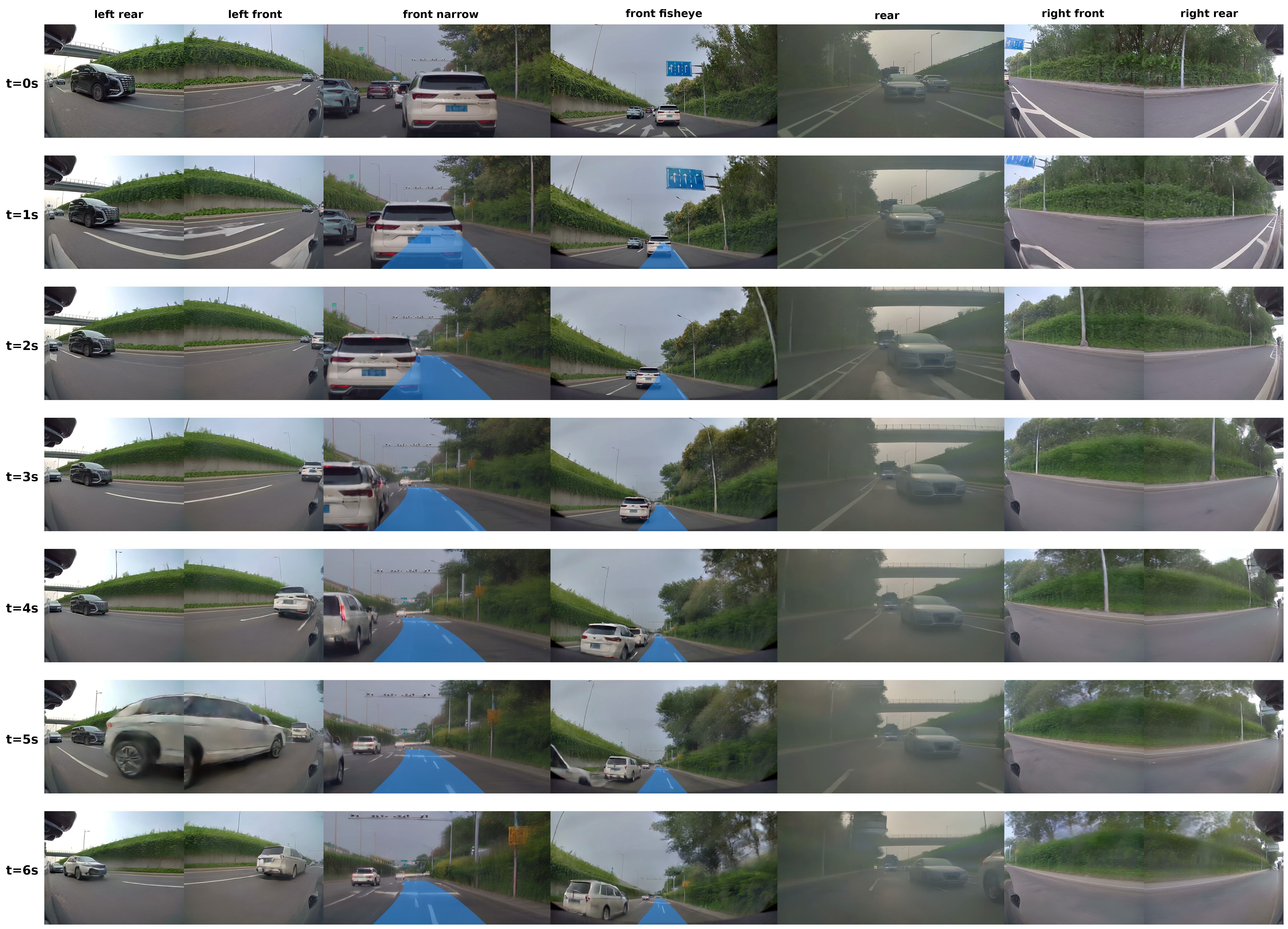}
  \caption{Visualization of the Vision Renderer conditioned on camera tokens.
  The horizontal axis spans the seven surround-view cameras (left rear,
  left front, front narrow, front fisheye, rear, right front, right rear),
  and the vertical axis spans time from $t=0$\,s (ground truth) to
  $t=6$\,s of predicted future. The blue overlay on the front-narrow and
  front-fisheye views shows the trajectory predicted by the LDM from action
  tokens, which the renderer never sees.}
  \label{fig:dit_vis}
\end{figure}
Fig~\ref{fig:dit_vis} visualized a 6-second rollout from the Vision Renderer across all seven surround-view cameras (left to right: left rear, left front, front narrow, front fisheye, rear, right front, right rear). The vertical axis was time, with the ground-truth observation at $t=0$\,s at the top and 6\,s of predicted future extending downward. Each AR step emitted one second of camera tokens at 4\,Hz, which the renderer decoded into four frames per camera; six consecutive steps yielded the full horizon shown.

To probe whether camera tokens encoded the ego-vehicle's intended motion, we overlaid on the front-narrow and front-fisheye views the trajectory predicted by the LDM from its action tokens (blue line). The renderer was not conditioned on these action tokens (Section~3), so the trajectory served as an independent reference: it revealed the LDM's intent but could not influence the rendered scene. Despite this separation, the trajectory remained tightly aligned with the synthesized frames across the full 6-second horizon, consistently tracing a path geometrically consistent with the lane configuration, surrounding agents, and ego pose. This indicated that the camera tokens already encoded the ego trajectory and scene dynamics, and that the renderer was genuinely controlled by the LDM's latent imagination rather than hallucinating from history alone.

Table~\ref{tab:renderer_metrics} reported quantitative image-quality metrics for both the Camera Latent Decoder and the Vision Renderer, evaluated at a 1-second horizon (single AR step) and at the 6-second horizon. FID~\cite{heusel2017gans} was computed over all seven surround-view cameras at 4\,Hz (four frames per camera per second), with ground-truth references sampled from 200k random clips in the test dataset. FVD~\cite{unterthiner2018towards} was computed per camera over its temporal sequence and then averaged across the seven cameras. The decoder reflected reconstruction quality of camera tokens in isolation, while the renderer measured end-to-end generation quality from the autoregressive rollout.
\begin{table}[h]
\centering
\caption{Image-quality metrics for the Camera Latent Decoder and Vision Renderer at 1\,s and 6\,s horizons.}
\label{tab:renderer_metrics}
\small
\begin{tabular}{l cccc}
\toprule
Method & \multicolumn{2}{c}{FID $\downarrow$} & \multicolumn{2}{c}{FVD $\downarrow$} \\
\cmidrule(lr){2-3} \cmidrule(lr){4-5}
& 1\,s & 6\,s & 1\,s & 6\,s \\
\midrule
Camera Latent Decoder & 10.97 & 11.82 & 135.56 & 158.39 \\
Vision Renderer       & 1.51  & 2.84  & 11.28  & 29.52  \\
\bottomrule
\end{tabular}
\end{table}

Taken together, Fig~\ref{fig:dit_vis} and
Table~\ref{tab:renderer_metrics} demonstrated that the Vision Renderer
produced high-fidelity multi-view imagery suitable for closed-loop
inference within \modelname. Quantitatively, the renderer reached
FID/FVD of $1.51/11.28$ at a 1\,s horizon and $2.84/29.52$ at 6\,s,
reducing FID/FVD relative to the Camera Latent Decoder by
$10.97/135.56$ at 1\,s and $11.82/158.39$ at 6\,s. The modest
degradation from 1\,s to 6\,s---an absolute increase of $1.33$ FID
points and $18.24$ FVD points---indicated that the autoregressive
rollout accumulated only limited drift across the full prediction
window, an essential property for closed-loop deployment where each
step's output conditions the next. Qualitatively, the LDM-predicted
trajectory in Fig~\ref{fig:dit_vis} remained geometrically consistent
with the rendered scene throughout the 6-second horizon despite the
renderer never receiving action tokens, confirming that the LDM's
camera tokens already encoded the ego trajectory and surrounding
scene dynamics with sufficient fidelity for the DiT to faithfully
visualize the LDM's latent imagination. Together, these results
established that the Vision Renderer met the fidelity,
controllability, and temporal-stability requirements needed to serve
as the pixel-space front-end of \modelname's closed-loop inference
loop.

\section{Conclusion}
\label{sec:conclusion}

We presented \modelname, a predictive world model that equipped VLA models with an internalized understanding of the physical world by forecasting high-fidelity multi-view camera observations while maintaining real-time action control. The proposed chunk-wise auto-regressive strategy addressed two key challenges in video-based world modeling: degeneration in future rollouts and the temporal mismatch between instantaneous dynamics and long-horizon world causality. Furthermore, we introduced a curriculum learning schedule that progressively extended prediction horizons, stabilizing training and improving control robustness. We also employed temporal importance sampling to concentrate supervision on safety-critical chunks, thereby enhancing the model’s ability to learn decision-relevant dynamics and better understand world causality. Finally, photorealistic synthesis was delegated to a diffusion-based multi-view renderer, which provided pixel-level supervision and improved rollout stability.

Our work extended world model research to industrial-scale data and established a foundation for several future research directions. First, incorporating additional supervision during closed-loop rollout might further improve long-tail performance in autonomous systems. Second, integrating richer modalities, such as 3D geometric supervision, might enhance the model’s physical world knowledge. Third, more advanced training strategies, such as coarse-to-fine world reconstruction, might help unlock the model’s full potential for comprehensive world understanding.
\newpage
\section*{Acknowledgments}
We sincerely thank our predictive world model team for their passionate exploration, valuable discussions, and hard work throughout this project. We also thank our sibling GWM team for their excellent work on X-World~\cite{zheng2026x}, which we build upon for pretraining our diffusion-based renderer. In particular, we are grateful to our advisors, Yu Zhang and Xianming Liu, for their insightful guidance on world model design, which was instrumental to the development of this work. We also thank XPeng for providing the platform, tools, and support that made this work possible.

We further acknowledge the broader research community, whose progress in world modeling, vision-language-action models, and generative modeling has inspired and informed our contribution to predictive world modeling.
\section*{Contributors}
\newcommand{\equalIT}{\textsuperscript{*}}
\newcommand{\internIT}{\textsuperscript{\ensuremath{\ddagger}}} 

\vspace{0.3em}

\begin{description}
    \item[Advisors:] Yu Zhang, Xianming Liu
    \item[Project Lead:] Zhuangzhuang Ding, Pengkun Zheng
    \item[Contributors:] Baolu Li\equalIT, Jingyu Qian\equalIT, Rui Guo\equalIT, Yilun Chen\equalIT, Hanpeng Liu, Yuan Lin, Junhong Zhou, Ruixin Liu, Liu Yang, Yutong Zheng, Zhenli Zhang, Sean Li, Chaoda Zheng, Boyang Wang
    \item[Technical Program Manager:] Tenglong (Victor) Gu

\end{description}

{\footnotesize
\noindent \equalIT Core contribution. The first four authors are listed in alphabetical order.}

% references
% \clearpage
{ 
\small
\bibliography{neurips_2025}

@String(cvpr  = {CVPR})

@String(iccv  = {ICCV})

@String(iclr  = {ICLR})

@String(icml  =	{ICML})

@article{brooks2024video,
  title={Video generation models as world simulators},
  author={Brooks, Tim and Peebles, Bill and Holmes, Connor and DePue, Will and Guo, Yufei and Jing, Leo and Schnurr, David and Taylor, Joe and Luhman, Troy and Luhman, Eric and others},
  journal={OpenAI Blog},
  volume={1},
  number={8},
  pages={1},
  year={2024}
}

@inproceedings{bruce2024genie,
  title={Genie: Generative interactive environments},
  author={Bruce, Jake and Dennis, Michael D and Edwards, Ashley and Parker-Holder, Jack and Shi, Yuge and Hughes, Edward and Lai, Matthew and Mavalankar, Aditi and Steigerwald, Richie and Apps, Chris and others},
  booktitle={International Conference on Machine Learning (ICML)},
  year={2024}
}

@article{zheng2026x,
  title={X-World: Controllable Ego-Centric Multi-Camera World Models for Scalable End-to-End Driving},
  author={Zheng, Chaoda and Li, Sean and Deng, Jinhao and Wang, Zhennan and Chen, Shijia and Xiao, Liqiang and Chi, Ziheng and Lin, Hongbin and Chen, Kangjie and Wang, Boyang and others},
  journal={arXiv:2603.19979},
  year={2026}
}

@inproceedings{assran2023self,
  title={Self-supervised learning from images with a joint-embedding predictive architecture},
  author={Assran, Mahmoud and Duval, Quentin and Misra, Ishan and Bojanowski, Piotr and Vincent, Pascal and Rabbat, Michael and LeCun, Yann and Ballas, Nicolas},
  booktitle={IEEE/CVF Conference on Computer Vision and Pattern Recognition (CVPR)},
  pages={15619--15629},
  year={2023}
}

@inproceedings{zitkovich2023rt,
  title={Rt-2: Vision-language-action models transfer web knowledge to robotic control},
  author={Zitkovich, Brianna and Yu, Tianhe and Xu, Sichun and Xu, Peng and Xiao, Ted and Xia, Fei and Wu, Jialin and Wohlhart, Paul and Welker, Stefan and Wahid, Ayzaan and others},
  booktitle={Conference on Robot Learning (CoRL)},
  pages={2165--2183},
  year={2023},
  organization={PMLR}
}

@article{driess2023palm,
  title={Palm-e: An embodied multimodal language model},
  author={Driess, Danny and Xia, Fei and Sajjadi, Mehdi SM and Lynch, Corey and Chowdhery, Aakanksha and Ichter, Brian and Wahid, Ayzaan and Tompson, Jonathan and Vuong, Quan and Yu, Tianhe and others},
  journal={arXiv:2303.03378},
  year={2023}
}

@article{kim2024openvla,
  title={Openvla: An open-source vision-language-action model},
  author={Kim, Moo Jin and Pertsch, Karl and Karamcheti, Siddharth and Xiao, Ted and Balakrishna, Ashwin and Nair, Suraj and Rafailov, Rafael and Foster, Ethan and Lam, Grace and Sanketi, Pannag and others},
  journal={arXiv:2406.09246},
  year={2024}
}

@inproceedings{liradial,
  title={Radial Attention: $\text{O}(n\log n) $ Sparse Attention with Energy Decay for Long Video Generation},
  author={Li, Xingyang and Li, Muyang and Cai, Tianle and Xi, Haocheng and Yang, Shuo and Lin, Yujun and Zhang, Lvmin and Yang, Songlin and Hu, Jinbo and Peng, Kelly and others},
  booktitle={Advances in Neural Information Processing Systems (NeurIPS)},
  year={2025}
  
}

@misc{guo2024blocksparse,
  author       = {Guo, Junxian and Tang, Haotian and Yang, Shang and Zhang, Zhekai and Liu, Zhijian and Han, Song},
  title        = {{Block Sparse Attention}},
  year         = {2024},
  publisher    = {GitHub},
  journal      = {GitHub repository},
  howpublished = {\url{https://github.com/mit-han-lab/Block-Sparse-Attention}}
}

@misc{xpeng2026vla2.0,
  author       = {{XPeng Inc.}},
  title        = {{XPeng VLA 2.0}},
  year         = {2026},
  howpublished = {\url{https://www.xpeng.com}}
}

@misc{world2026marble,
  author       = {World Labs},
  title        = {{Marble: World Labs Spatial Intelligence}},
  year         = {2026},
  publisher    = {World Labs},
  journal      = {World Labs Website},
  howpublished = {\url{https://www.worldlabs.ai/}}
}

@inproceedings{peebles2023scalable,
  title={Scalable diffusion models with transformers},
  author={Peebles, William and Xie, Saining},
  booktitle={IEEE/CVF International Conference on Computer Vision (ICCV)},
  pages={4195--4205},
  year={2023}
}

@article{shi2025,
  title={{SVG}: Latent Diffusion Model without Variational Autoencoder},
  author={Minglei Shi and Haolin Wang and Wenzhao Zheng and Ziyang Yuan and Xiaoshi Wu and Xintao Wang and Pengfei Wan and Jie Zhou and Jiwen Lu},
  journal={arXiv:2510.15301},
  year={2025}
}

@inproceedings{lipman2023flow,
  title     = {Flow Matching for Generative Modeling},
  author    = {Lipman, Yaron and Chen, Ricky T. Q. and Ben-Hamu, Heli and
               Nickel, Maximilian and Le, Matt},
  booktitle = {International Conference on Learning Representations (ICLR)},
  year      = {2023},
  url       = {https://openreview.net/forum?id=PqvMRDCJT9t}
}

@article{wan2025wan,
  title={Wan: Open and advanced large-scale video generative models},
  author={Wan, Team and Wang, Ang and Ai, Baole and Wen, Bin and Mao, Chaojie and Xie, Chen-Wei and Chen, Di and Yu, Feiwu and Zhao, Haiming and Yang, Jianxiao and others},
  journal={arXiv:2503.20314},
  year={2025}
}

@inproceedings{yin2025slow,
  title={From slow bidirectional to fast autoregressive video diffusion models},
  author={Yin, Tianwei and Zhang, Qiang and Zhang, Richard and Freeman, William T and Durand, Fredo and Shechtman, Eli and Huang, Xun},
  booktitle={IEEE/CVF Conference on Computer Vision and Pattern Recognition (CVPR)},
  pages={22963--22974},
  year={2025}
}

@article{huang2026self,
  title={Self forcing: Bridging the train-test gap in autoregressive video diffusion},
  author={Huang, Xun and Li, Zhengqi and He, Guande and Zhou, Mingyuan and Shechtman, Eli},
  journal={Advances in Neural Information Processing Systems (NeurIPS)},
  volume={38},
  pages={167283--167308},
  year={2026}
}

@article{yuan2026helios,
  title={Helios: Real Real-Time Long Video Generation Model},
  author={Yuan, Shenghai and Yin, Yuanyang and Li, Zongjian and Huang, Xinwei and Yang, Xiao and Yuan, Li},
  journal={arXiv:2603.04379},
  year={2026}
}

@article{dosovitskiy2020image,
  title={An image is worth 16x16 words: Transformers for image recognition at scale},
  author={Dosovitskiy, Alexey and Beyer, Lucas and Kolesnikov, Alexander and Weissenborn, Dirk and Zhai, Xiaohua and Unterthiner, Thomas and Dehghani, Mostafa and Minderer, Matthias and Heigold, Georg and Gelly, Sylvain and others},
  journal={arXiv:2010.11929},
  year={2020}
}

@inproceedings{dao2024flashattention,
  title={Flashattention-2: Faster attention with better parallelism and work partitioning},
  author={Dao, Tri},
  booktitle={International Conference on Learning Representations},
  volume={2024},
  pages={35549--35562},
  year={2024}
}

@article{Gao2025OneLI,
  title={One Layer Is Enough: Adapting Pretrained Visual Encoders for Image Generation},
  author={Yuan Gao and Chen Chen and Tianrong Chen and Jiatao Gu},
  journal={ArXiv},
  year={2025},
  volume={2512.07829},
}

@article{heusel2017gans,
  title={Gans trained by a two time-scale update rule converge to a local nash equilibrium},
  author={Heusel, Martin and Ramsauer, Hubert and Unterthiner, Thomas and Nessler, Bernhard and Hochreiter, Sepp},
  journal={Advances in Neural Information Processing Systems (NeurIPS)},
  volume={30},
  year={2017}
}

@article{unterthiner2018towards,
  title   = {Towards Accurate Generative Models of Video: A New Metric \& Challenges},
  author  = {Unterthiner, Thomas and van Steenkiste, Sjoerd and Kurach, Karol and Marinier, Rapha{\"e}l and Michalski, Marcin and Gelly, Sylvain},
  journal = {arXiv:1812.01717},
  year    = {2018}
}
\bibliographystyle{plain} 
}

\end{document}